\documentclass[11pt]{article}
\usepackage{times}
\usepackage{sibgrapi}
\usepackage{graphicx}
\usepackage{amssymb}
\usepackage{amsmath}

\DeclareGraphicsExtensions{.jpg, .pdf, .mps, .png}

\input{epsf}        

\DeclareGraphicsExtensions{.jpg, .pdf, .mps, .png}

\begin{document}

\title{Fourier Analysis \\
            and \\
            Holographic Representations of $1D$ and $2D$ Signals}
\author{G.A. Giraldi\authortag{1} and  B.F. Moutinho\authortag{1,2} \\
              and \\
         D.M.L. de Carvalho\authortag{3} and J.C. de Oliveira\authortag{1}}



\address{
  \authortag{1}LNCC--National Laboratory for Scientific Computing -\\
  Av. Getulio Vargas, 333, 25651-070, Petr\'opolis, RJ, Brazil\\
  {\tt \{pssr,gilson\}@lncc.br}\\
 \authortag{2}Catholic University of Petr\'opolis \\
              Engineering Department - Petr\'opolis, RJ, Brazil \\
  {\tt brunofig@lncc.br}\\
   \authortag{3} Military Institute of Engineering  \\
                 Rio de Janeiro, RJ, Brazil \\
  {\tt dulio@centroin.com.br}
}

\abstract{\ In this paper, we focus on Fourier analysis and
holographic transforms for signal representation. For instance, in
the case of image processing, the holographic representation has the
property that an arbitrary portion of the transformed image enables
reconstruction of the whole image with details missing. We focus on
holographic representation defined through the Fourier Transforms.
Thus, We firstly review some results in Fourier transform and
Fourier series. Next, we review the Discrete Holographic Fourier
Transform (\textbf{DHFT}) for image representation. Then, we
describe the contributions of our work. We show a simple scheme for
progressive transmission based on the DHFT. Next, we propose the
Continuous Holographic Fourier Transform (\textbf{CHFT}) and discuss
some theoretical aspects of it for $1D$ signals.}

\DeclareGraphicsExtensions{.jpg, .pdf, .mps, .png}

\maketitle

\section{Introduction}

The techniques for signal representation composes a class of methods
used to characterize the information therein
\cite{Gonzalez1992,Jain89}. In the area of digital image processing,
a classical approach for image representation is based on unitary
matrices that define the so called \textit{image transforms.} The
Discrete Fourier Transform (\textbf{DFT}) is the most known example
in this class.

Stochastic models describe an image as a member of an ensemble,
which can be characterized by its mean and covariance functions.
This allows development of algorithms that are useful for an entire
class or an ensemble of images rather than for a single image. Fall
in this category the covariance models, the $1-D$ models (an image
is considered a one-dimensional random signal that appears at the
output of a raster scanner) and $2-D$ models (Causal, Semicausal and
Noncausal) \cite{Jain89}.

Multiresolution representations, such Pyramid \cite{Montanvert92}
and Wavelets \cite{Zeevi1998}, compose another important class in
this field. Finally, for image understanding models, an image can be
considered as a composition of several objects detected by
segmentation and computer vision techniques
\cite{Gonzalez1992,Jain89}.

In this paper we focus on the first category of image representation
methods. We firstly review some results in Fourier transform and
Fourier series and show some connections between them (section
\ref{FourierAnalysis}). Next, in section \ref{sec:DHFT}, we describe
the discrete Holographic Fourier Transform (\textbf{DHFT}) proposed
in \cite {Netravali1998}, which is based on the DFT. The holographic
representation of an image has the property that an arbitrary
portion of the holographic representation enables reconstruction of
the whole image with details missing. This is an interesting
property that motivates works using holographic representations for
image compression \cite{877155}, multimedia systems
\cite{Ferri-1996}, watermarking (see \cite{Netravali1998} and
references therein) and representations of line images
\cite{DBLP:journals/prl/Gorsky94}. Besides the holographic
representation based on the DFT there are also approaches that use
subsampling methods \cite{841469}.

In sections \ref{CHT} and \ref{ProgT}, we show our contributions in
this field. We propose the Continuous Holographic Fourier Transform
(section \ref{CHT}) and discuss some theoretical and numerical
aspect of it. Then, in section \ref{ProgT}, we develop a simple
scheme for progressive transmission of images through the DHFT.
 Section \ref{Concl} presents
our conclusions and some comments. The Appendix reviews interesting
results in discrete Fourier analysis. An extended version of this
work can be found in \cite{Gilson-BrunoFig2006}

\section{Fourier Analysis Revised \label{FourierAnalysis}}

In this section we review some important results about continuous
Fourier Analysis. A special attention will be given to the
connections between the Fourier Transform and Fourier Series
representation of a signal. Thus, let us begin with usual
definitions in this area. For simplicity, we restrict our discussion
to the one dimensional case. Generalizations to higher-dimensional
cases are straightforward. In this section our presentation follows
the reference \cite{Chui1992}.

Throughout this text, all functions $f$ are defined on the real line, with
values in the complex domain $C$, $f:\Re \rightarrow C$. Besides, they are
assumed to be piecewise continuous and having the integral:
\begin{equation}
\int_{-\infty }^{\infty }\left\vert f\left( x\right) \right\vert ^{p}dx,
\label{fourier00}
\end{equation}
finite. The set of such functions is denoted by $L^{P}\left( \Re \right) $.
In this set, the $p$-norm and the inner product are well defined by:

\begin{equation}
\left\Vert f\right\Vert _{p}=\left\{
\int_{-\infty}^{\infty}\left|f\left(x\right)\right|^{p}dx\right\}
^{1/p},\quad1\leq p<\infty,  \label{fourier01}
\end{equation}

\begin{equation}
\left\langle f,g\right\rangle =\int_{-\infty}^{\infty}f\left(x\right)%
\overline{g\left(x\right)}dx,\quad f,g\in L^{2}\left(\Re\right).
\label{fourier02}
\end{equation}

\textit{Definition 1:} Given a function $f\in L^{1}\left(\Re\right),$ its
Fourier Transform is defined by:

\begin{equation}
\widehat{f}\left(\omega\right)=\left(\digamma
f\right)\left(\omega\right)=\int_{-\infty}^{\infty}f\left(x\right)\exp%
\left(-2\pi j\omega x\right)dx.  \label{fourier03}
\end{equation}

\textit{Definition 2:} Given the Fourier transform $\widehat{f}$, the
transformed function $f$ can be obtained by the Inverse Fourier Transform,
given as follows:

\begin{equation}
f\left(x\right)=\left(\digamma^{-1}\widehat{f}\right)\left(x\right)=\int_{-%
\infty}^{\infty}\widehat{f}\left(\omega\right)\exp\left(2\pi
jx\omega\right)d\omega.  \label{inverse00}
\end{equation}

\textit{Definition 3:} Let $f$ and $g$ be functions in $L^{1}\left( \Re
\right) $. Then the \textit{convolution} of $f$ and $g$ is also an $%
L^{1}\left( \Re \right) $ function $h$ which is defined by:

\begin{equation}
h\left( x\right) =\left( f*g\right) \left( x\right) =\int_{-\infty }^{\infty
}f\left( x-y\right) g\left( y\right) dy.  \label{fourier05}
\end{equation}

\textit{Theorem 1:} Any function $f\in L^{2}\left( -L/2,L/2\right) $ has a
Fourier series representation given by:

\begin{equation}
f\left( x\right) =\sum_{n=-\infty }^{+\infty }a_{n}\exp \left( 2j\pi x\frac{n%
}{L}\right) .  \label{fourier07}
\end{equation}

Dem: See \cite{DjairoIMPA}.

Among the properties of the Fourier Transform, the following ones
have a special place for signal processing techniques, according to our discussion
presented in \cite{Gilson-BrunoFig2006}.

\textit{Property 1:} If the derivative $df/dx$ of $f$ exists and is in $%
L^{1}\left(\Re\right)$, then:

\begin{equation}
\left[\digamma\left(\frac{df}{dx}\right)\right]\left(\omega\right)=\widehat{%
f^{\prime}}\left(\omega\right)=j\omega\widehat{f}.  \label{fourier04}
\end{equation}

\textit{Property 2:} $\widehat{f}\rightarrow 0,$ as $\omega \rightarrow
\infty $ or $\omega \rightarrow -\infty .$

For the proofs, see \cite{Chui1992}, pp. 25.

\textit{Property 3 (Convolution Theorem):} The Fourier transform of the
convolution of two functions is the product of their Fourier transforms,
that is:

\begin{equation}
h\left( x\right) =\left( f\ast g\right) \left( x\right) \Leftrightarrow
\widehat{h}\left( \omega \right) =\widehat{f}\left( \omega \right) \widehat{g%
}\left( \omega \right) .  \label{fourier06}
\end{equation}

Dem: See \cite{DjairoIMPA}.

An important aspect for signal (and image) processing is the relationship
between the coefficients $a_{n}$ in the series (\ref{fourier07}) and the
Fourier transform defined by the integral (\ref{fourier03}), in the case of
a function $f\in L^{2}\left( -L/2,L/2\right) $. In order to find this
relationship, let us remember that, from the orthogonality of the functions $%
\left\{ \exp \left( 2j\pi x\frac{n}{L}\right) ,n\in Z\right\} ,$ it
is easy to show what the coefficients $a_{n}$ can be obtained by the
expression:

\begin{equation}
a_{n}=\frac{1}{L}\int_{-L/2}^{L/2}f\left( x\right) \exp \left( -2\pi jx\frac{%
n}{L}\right) dx.  \label{fourier08}
\end{equation}

Now, we must observe that this expression is equivalent to:

\begin{equation}
\frac{1}{L}\widehat{f}\left(\frac{n}{L}\right)=\frac{1}{L}%
\int_{-\infty}^{\infty}f\left(x\right)\exp\left(-2\pi jx\frac{n}{L}\right)dx,
\label{fourier09a}
\end{equation}
once $f\in L^{2}\left(-L/2,L/2\right).$\newline

Thus,

\begin{equation}
a_{n}=\frac{1}{L}\widehat{f}\left( \frac{n}{L}\right) .  \label{fourier09b}
\end{equation}

This result shows the connections between Fourier series and the Fourier
Transform. It is used in \cite{Gilson-BrunoFig2006} to discussion Fourier analysis in te context of
signal processing.

\section{Discrete Holographic Fourier Transform \label{sec:DHFT}}

Optical holography technology uses interference and diffraction of
light to record and reproduce $3D$ information of an optical
wavefront \cite{Balogh:2005:SHS,Tim-Chuang2006}. This technology has
been applied for $3D$ displays \cite{364111} and data storage
devices \cite{DBLP:ibm/journal/research/}. From an arbitrary portion
of an optical hologram that encodes a scene, a low quality version
of the entire scene can be reconstructed. The quality of the
recovered signal depends only on the size of the hologram portion
used, but not of the place from where it was taken
\cite{Image-Formation}. The holographic representation of images
tries to mimic such property through image transforms. In this
section, we summarize the results presented in \cite{Netravali1998}.

The Discrete Holographic Fourier Transform (\textbf{DHFT}) of an image $%
I=I\left( x,y\right) $ is defined as \cite{Netravali1998}:

\begin{equation}
H\left( u,v\right) =\digamma ^{-1}\left[ I\left( x,y\right) e^{2\pi jP\left(
x,y\right) }\right] ,  \label{holo00}
\end{equation}
where $P\left( x,y\right) $ is a random phase image so that $E\left[
P\left( x,y\right) P\left( \overline{x},\overline{y}\right) \right]
=0$ for $\left( x,y\right) \neq \left(
\overline{x},\overline{y}\right) $ and $P\left(
x,y\right) $ is a random variable uniformly distributed over $\left[ -1,1%
\right] .$ By holographic representation, we mean that from an image portion
$H^{c}\left( u,v\right) $ cropped from anywhere in the complex image $%
H\left( u,v\right) ,$ we can, by $2D$ Fourier transformation, get a version
of $I\left( x,y\right) $ so that the degradation is proportional to the size
of $H^{c}\left( u,v\right) $. Clearly, if $H\left( u,v\right) $ is available
we get back $I\left( x,y\right) $ as the amplitude of its $2D$ Fourier
transform, and the same image recovery process will be applied over cropped
parts of $H\left( u,v\right) $. Certainly, we can take a portion of an
unitary transform and perform the same process. However the result will be
strong dependent from the place from where the portion was cropped.

In order to demonstrate that the image transform defined in
expression (\ref{holo00}) has the holographic property, let us
follow the development found in \cite{Netravali1998} and consider
the $1D$ discrete version of the above proposed representation
method. The one-dimensional DHFT is:

\begin{equation}
H\left(u\right)=\sum\limits _{k=0}^{M-1}I\left(k\right)e^{j2\pi
P\left(k\right)}\frac{1}{\sqrt{M}}e^{j\frac{2\pi}{M}uk},  \label{holo01}
\end{equation}
where $\left\{ P\left(k\right)\right\} $ is a set of independent and
identically distributed random numbers uniformly distributed over $\left[0,1%
\right]$. We will represent the process of cropping a portion of $%
H\left(u\right)$ by multiplying it with a window function $W\left(u\right)$,
given by:

\begin{equation}
W\left( u\right) =\left\{
\begin{array}{c}
1, u \in \left[ a,a+\left( L-1\right) \right], \\
0, u \notin \left[ a,a+\left( L-1\right) \right],
\end{array}
\right.  \label{win00}
\end{equation}
where $a\in \left\{ 0,1,...,M-L\right\} $ for simplicity. Now, we
shall consider the question: what can be recovered from $H^{c}\left(
u\right) =H\left( u\right) \cdot W\left( u\right) $ by the
one-dimensional Fourier transform. To answer this question we call
$I_{W}\left( r\right) $ the obtained signal. Thus:

\begin{equation*}
I_{w}\left(r\right)=\sum\limits _{u=0}^{M-1}H\left(u\right)W\left(u\right)%
\frac{1}{\sqrt{M}}e^{-j\frac{2\pi}{M}ur}=
\end{equation*}

\begin{equation*}
\sum\limits _{u=0}^{M-1}\left(\sum\limits
_{k=0}^{M-1}I\left(k\right)e^{j2\pi P\left(k\right)}\frac{1}{\sqrt{M}}e^{j%
\frac{2\pi}{M}uk}\right)W\left(u\right)\frac{1}{\sqrt{M}}e^{-j\frac{2\pi}{M}%
ur}=
\end{equation*}

\begin{equation*}
\frac{1}{M}\sum\limits _{u=0}^{M-1}\left(\sum\limits
_{k=0}^{M-1}I\left(k\right)e^{j2\pi P\left(k\right)}e^{j\frac{2\pi}{M}%
u\left(k-r\right)}W\left(u\right)\right)=
\end{equation*}

\begin{equation}
\frac{1}{M}\sum\limits_{k=0}^{M-1}\left( \sum\limits_{u=0}^{M-1}e^{-j\frac{%
2\pi }{M}u\left( r-k\right) }W\left( u\right) \right) I\left( k\right)
e^{j2\pi P\left( k\right) }.  \label{holo001}
\end{equation}

Let us define $g\left( r,k\right) $ as follows:

\begin{equation}
g\left( r,k\right) =\sum\limits_{u=0}^{M-1}e^{-j\frac{2\pi }{M}u\left(
r-k\right) }W\left( u\right) =\sum\limits_{u=a}^{a+(L-1)}e^{-j\frac{2\pi }{M}%
u\left( r-k\right) }.  \label{holo001a}
\end{equation}

Thus, by changing variable $\left( \alpha =u-a\right) $, we find:

\begin{equation}
g\left( r,k\right) =e^{-j\frac{2\pi }{M}\left( r-k\right)
a}\sum\limits_{\alpha =0}^{L-1}e^{-j\frac{2\pi }{M}\alpha \left( r-k\right)
}.  \label{holo001b}
\end{equation}

But, from the Property A2, in the Appendix:

\begin{equation}
\sum\limits_{u=0}^{L-1}e^{-j2\pi xu}=Le^{-j\left(L-1\right)\pi x}\frac{sinc
L\pi x}{sinc \pi x}.  \label{holo04}
\end{equation}

Through expressions (\ref{holo04}) and (\ref{holo001b}) we can
rewrite equation (\ref{holo001}) as:

\begin{equation*}
I_{W}\left(r\right)=\frac{1}{M}\sum\limits _{k=0}^{M-1}e^{-j\frac{2\pi}{M}%
\left(r-k\right)a}e^{-j\left(L-1\right)\pi\frac{\left(r-k\right)}{M}}\frac{%
Lsinc \frac{L\pi\left(r-k\right)}{M}}{sinc \frac{\pi\left(r-k\right)}{M}}%
I\left(k\right)e^{j2\pi P\left(k\right)}.
\end{equation*}

Thus, we finally have:

\begin{equation*}
I_{w}\left(r\right)=\sum\limits _{k=0}^{M-1}e^{-j\frac{2\pi}{M}ra}e^{j\frac{%
2\pi}{M}ka}e^{-j\left(L-1\right)\frac{\pi}{M}r}e^{j\left(L-1\right)\frac{\pi%
}{M}k}\frac{L sinc \frac{L\pi\left(r-k\right)}{M}}{M sinc \frac{%
\pi\left(r-k\right)}{M}}I\left(k\right)e^{j2\pi P\left(k\right)}=
\end{equation*}

\begin{equation*}
=e^{-j\frac{2\pi }{M}\left[ a+\frac{\left( L-1\right) }{2}\right] r}\left(
\sum\limits_{k=0}^{M-1}I\left( k\right) e^{j2\pi \left[ P\left( k\right) +%
\frac{k}{M}\left( a+\frac{L-1}{2}\right) \right] }\frac{L sinc \frac{L\pi
\left( r-k\right) }{M}}{M sinc \frac{\pi \left( r-k\right) }{M}}\right) =
\end{equation*}

\begin{equation}
=e^{-j\frac{2\pi }{M}\left[ a+\frac{\left( L-1\right) }{2}\right] r}\left(
\sum\limits_{k=0}^{M-1}I\left( k\right) e^{j2\pi \tilde{P_{a}}\left(
k\right) }\frac{L sinc \frac{L\pi \left( r-k\right) }{M}}{M sinc \frac{\pi
\left( r-k\right) }{M}}\right).  \label{holo05}
\end{equation}

Let us define:

\begin{equation}
\phi _{L}\left( r-k\right) :=\frac{L sinc \frac{L\pi \left( r-k\right) }{M}}{%
M sinc \frac{\pi \left( r-k\right) }{M}}.
\end{equation}

Thus, we have:

\begin{equation}
I_{W}\left( r\right) =e^{-j\frac{2\pi }{M}\left[ a+\frac{\left( L-1\right) }{%
2}\right] r}\left( \sum\limits_{k=0}^{M-1}I\left( k\right) e^{j2\pi \tilde{%
P_{a}}\left( k\right) }\phi _{L}\left( r-k\right) \right).  \label{holo05a}
\end{equation}

Firstly, we should verify the result when $L=M$, and consequently,
$a=0$. In this case, expression (\ref{holo05}) becomes:

\begin{equation}
I_{W}\left( r\right) =e^{-j\frac{2\pi }{M}\left( \frac{M-1}{2}\right)
r}\sum\limits_{k=0}^{M-1}I\left( k\right) e^{j2\pi \left[ P\left( k\right) +%
\frac{k\left( M-1\right) }{2M}\right] }\phi _{L}\left( r-k\right) .
\label{holo05b}
\end{equation}

Besides, from the definition of $\phi _{L}$ we can see that, when
$L=M$, we have:

\begin{equation}
\phi _{L}\left( r-k\right) =\left\{
\begin{array}{l}
1, r-k=0, \\
0, r-k\neq 0,
\end{array}
\right.  \label{holo05c}
\end{equation}
because $0\leq r,k\leq M-1.$ Finally, by using the result
(\ref{holo05c}) in expression (\ref{holo05b}) we find:

\begin{equation*}
I_{W}\left( r\right) =e^{-j\frac{2\pi }{M}\left( \frac{M-1}{2}\right)
r}\sum\limits_{k=0}^{M-1}I\left( k\right) e^{j2\pi \left[ P\left( k\right) +%
\frac{k\left( M-1\right) }{2M}\right] }\phi _{L}\left( r-k\right) =
\end{equation*}

\begin{equation*}
e^{-j\frac{2\pi }{M}\left( \frac{M-1}{2}\right) r}I\left( r\right) e^{j2\pi %
\left[ P\left( r\right) +\frac{r\left( M-1\right) }{2M}\right] }=
\end{equation*}

\begin{equation*}
=e^{-j\frac{2\pi }{M}\left( \frac{M+1}{2}\right) r}e^{j\frac{2\pi }{M}\left(
\frac{M+1}{2}\right) r}I\left( r\right) e^{2\pi P\left( r\right) }=  \notag
\end{equation*}

\begin{equation}
=I\left( r\right) e^{j2\pi P\left( r\right)}.  \label{image000}
\end{equation}

So, as expected, the whole image is recovered as the amplitude of
the above result. In order to know how good will be the
approximation of the original image generated when $L<M$, we can
perform a statistical analysis of the recovered complex image.

Following \cite{Netravali1998}, we will assume that $I\left(
k\right) =I_{0}$ constant:

\begin{equation*}
I_{W}\left( r\right) =e^{-j\frac{2\pi }{M}\left[ a+\frac{\left( L-1\right) }{%
2}\right] r}\left( \sum\limits_{k=0}^{M-1}I\left( k\right) e^{j2\pi \left[
P\left( k\right) +\frac{k}{M}\left( a+\frac{L-1}{2}\right) \right] }\phi
_{L}\left( r-k\right) \right)
\end{equation*}

\begin{equation*}
=I_{0}\sum\limits _{k=0}^{M-1}e^{j2\pi\left[P\left(k\right)+\frac{%
\left(k-r\right)}{M}\left(a+\frac{L-1}{2}\right)\right]}\phi_{L}\left(r-k%
\right)=
\end{equation*}

\begin{equation*}
=I_{0}\sum\limits_{k=0}^{M-1}e^{j2\pi \tilde{P_{a}}\left( k-r\right) }\phi
_{L}\left( r-k\right) .
\end{equation*}

Henceforth:

\begin{equation*}
\left| I_{W}\left( r\right) \right| ^{2}=I_{W}\left( r\right) I_{W}\left(
r\right) ^{*}=
\end{equation*}

\begin{equation*}
=\left| I_{0}\right| ^{2}\left( \sum\limits_{k=0}^{M-1}e^{j2\pi \tilde{P_{a}}%
\left( k-r\right) }\phi _{L}\left( r-k\right) \right) \left(
\sum\limits_{l=0}^{M-1}e^{-j2\pi \tilde{P_{a}}\left( l-r\right) }\phi
_{L}\left( r-l\right) \right) =
\end{equation*}

\begin{equation*}
=\left| I_{0}\right| ^{2}\left\{ \sum\limits_{k=0}^{M-1}\phi _{L}^{2}\left(
r-k\right) +\sum\limits_{k\neq l}\sum\limits_{k,l=0}^{M-1}e^{j2\pi \tilde{%
P_{a}}\left( k-r\right) }\phi _{L}\left( r-k\right) e^{-j2\pi \tilde{P_{a}}%
\left( l-r\right) }\phi _{L}\left( r-l\right) \right\} =
\end{equation*}

\begin{equation*}
=\left| I_{0}\right| ^{2}\left\{ \sum\limits_{k=0}^{M-1}\phi _{L}^{2}\left(
r-k\right) +\sum\limits_{k\neq l}\sum\limits_{k,l=0}^{M-1}e^{j2\pi \left[
\tilde{P_{a}}\left( k-r\right) -\tilde{P_{a}}\left( l-r\right) \right] }\phi
_{L}\left( r-k\right) \phi _{L}\left( r-l\right) \right\} .
\end{equation*}

To answer how good an estimate is $\left| I_{W}\left( r\right)
\right| $ for the value $I_{0}$, it is used in \cite{Netravali1998}
the following result:

\textit{Lemma 1:} Let $\theta _{0},\theta _{1},...,\theta _{M-1}$ be
independent and identically distributed random numbers uniformly
chosen from $\left[ 0,1\right] $, and the random variable:

\begin{equation*}
V=\sum\limits_{k=0}^{M-1}e^{j2\pi \theta _{k}}\varphi \left(
k\right) , \label{lemma000new}
\end{equation*}
where $\varphi \left( k\right)$ is a sequence of real numbers. Thus,
we have the following statistics for $V$:

\begin{equation}
E\left( V\right) =0,\quad E\left( \left| V^{2}\right| \right)
=\sum\limits_{k=0}^{M-1}\varphi ^{2}\left( k\right) ,  \label{lemma001}
\end{equation}

\begin{equation}
\sigma \left( \left| V^{2}\right| \right) =\sqrt{\sum\limits_{k\neq
l}\sum\limits_{k,l=0}^{M-1}}\varphi ^{2}\left( k\right) \varphi ^{2}\left(
l\right) .  \label{lema002}
\end{equation}

As $I_{W}\left( r\right)$ is of the
form$\sum\limits_{k=0}^{M-1}e^{j2\pi \theta _{k}}\varphi \left(
k\right) $, it follows from this Lemma that:

\begin{equation*}
E\left[ \left| I_{W}\left( r\right) \right| ^{2}\right] =\left| I_{0}\right|
^{2}\sum\limits_{k=0}^{M-1}\phi _{L}^{2}\left( r-k\right) ,
\end{equation*}

\begin{equation*}
\sigma \left[ \left| I_{W}\left( r\right) \right| ^{2}\right] =\left|
I_{0}\right| ^{2}\sqrt{\sum\limits_{k\neq l}\sum\limits_{k,l=0}^{M-1}\phi
_{L}^{2}\left( r-k\right) \phi _{L}^{2}\left( r-l\right) }.
\end{equation*}

If $L=M$, expression (\ref{holo05c}) implies that:

\begin{equation*}
E\left[ \left| I_{W}\left( r\right) \right| ^{2}\right] =\left| I_{0}\right|
^{2},\quad \sigma \left[ \left| I_{W}\left( r\right) \right| ^{2}\right] =0,
\end{equation*}
as expected from the result (\ref{image000}).

Supposing $L<M$, it follows:

\begin{equation*}
E\left[ \left| I_{W}\left( r\right) \right| ^{2}\right] =\left| I_{0}\right|
^{2}\sum\limits_{k=0}^{M-1}\phi _{L}^{2}\left( r-k\right) =
\end{equation*}

\begin{equation*}
=\left| I_{0}\right| ^{2}\sum\limits_{k=0}^{M-1}\frac{L^{2} sinc^{2}\frac{%
L\pi \left( r-k\right) }{M}}{M^{2} sinc^{2}\frac{\pi \left( r-k\right) }{M}}=
\end{equation*}

\begin{equation}
=\frac{L^{2}}{M^{2}}\left| I_{0}\right| ^{2}\sum\limits_{k=0}^{M-1}\frac{%
sinc^{2}\frac{L\pi \left( r-k\right) }{M}}{sinc^{2}\frac{\pi \left(
r-k\right) }{M}}.  \label{last-exp}
\end{equation}

Simplifying the summation:

\begin{equation*}
\sum\limits_{k=0}^{M-1}\frac{sinc^{2}\frac{L\pi \left( r-k\right) }{M}}{%
sinc^{2}\frac{\pi \left( r-k\right) }{M}}=\frac{M}{L},
\end{equation*}
so, returning to expression (\ref{last-exp}) we find:

\begin{equation*}
E\left[\left|I_{W}\left(r\right)\right|^{2}\right]=\frac{L^{2}}{M^{2}}%
\left.I_{0}\right.^{2}\sum\limits _{k=0}^{M-1}\frac{sinc^{2}\frac{%
L\pi\left(r-k\right)}{M}}{sinc^{2}\frac{\pi\left(r-k\right)}{M}}=\frac{L}{M}%
\left.I_{0}\right.^{2},
\end{equation*}

\begin{equation*}
E\left[\left|I_{W}\left(r\right)\right|^{2}\right]=\frac{L}{M}%
\left.I_{0}\right.^{2}.
\end{equation*}

So the image recovered is an approximation of the original image
multiplied by the factor $\sqrt{\frac{L}{M}}.$

In the previous demonstrations we recovered the original image from
a window in the holographic domain extended by zeros. We consider
now recovering the image from the window without extending it. The
Fourier transform of the cropped sequence $H(u)$, $u\in \left[
a,a+\left( L-1\right) \right] $, is:

\begin{equation}
I_{T}\left( r\right) =FT\left[ H\left( u\right) \right] =\sum%
\limits_{k=0}^{L-1}H\left( k+a\right) \frac{1}{\sqrt{L}}e^{-j\frac{2\pi }{L}%
rk}.  \label{fseq00}
\end{equation}

Defining the extended sequence $\tilde{H}(u)$, $u\in \left[
0,M-1\right] $ and its Fourier transform as below:

\begin{equation*}
\tilde{H}(k)=\left\{
\begin{array}{l}
H\left( k+a\right) ,\quad k\in \left[ 0,\quad L-1\right], \\
0, \quad k\in \left[ L,\quad M-L\right],
\end{array}
\right.
\end{equation*}

\begin{equation*}
I_{WS}\left( v\right) =FT\left[ \tilde{H}(u)\right] =\sum\limits_{u=0}^{M-1}%
\tilde{H}(u)\frac{1}{\sqrt{M}}e^{-j\frac{2\pi }{M}vu}=
\end{equation*}

\begin{equation}
=\sum\limits_{k=0}^{L-1}H\left( k+a\right)
\frac{1}{\sqrt{M}}e^{-j\frac{2\pi }{M}vk}.  \label{fseq01}
\end{equation}

Comparing the two expressions (\ref{fseq00}) and (\ref{fseq01}) it
can be straightforward noticed that:

\begin{equation}
I_{WS}\left( v\right)=\sqrt{\frac{L}{M}}I_{T}\left( \frac{L}{M}%
v\right),\quad v=0,\quad \frac{L}{M},\quad \frac{2L}{M},\quad ...,\quad
\frac{\left( L-1\right) L}{M}.  \label{fseq01a}
\end{equation}

We must also observe that:

\begin{equation}
I_{w}\left( r\right) =\sum\limits_{u=a}^{a+L-1}H\left( u\right) \frac{1}{%
\sqrt{M}}e^{-j\frac{2\pi }{M}ur}.
\end{equation}

Hence, by changing $k=u-a$, we have

\begin{equation*}
I_{w}\left( r\right) =\sum\limits_{k=0}^{L-1}H\left( k+a\right) \frac{1}{%
\sqrt{M}}e^{-j\frac{2\pi }{M}(k+a)r}=
\end{equation*}

\begin{equation}
=\left( \sum\limits_{k=0}^{L-1}H\left( k+a\right) \frac{1}{\sqrt{M}}e^{-j%
\frac{2\pi }{M}kr}\right) e^{-j\frac{2\pi }{M}ar}=I_{WS}\left( r\right)
\cdot e^{-j\frac{2\pi }{M}ar}.  \label{fseq01b}
\end{equation}

\begin{equation}
I_{WS}\left( r\right) =I_{W}\left( r\right) \cdot e^{j\frac{2\pi }{M}ar}.
\label{fseq01c}
\end{equation}

Substituting this result in (\ref{fseq01a}), we have:

\begin{equation}
I_{T}\left( \frac{L}{M}v\right) =\sqrt{\frac{M}{L}}I_{W}\left( v\right) e^{j%
\frac{2\pi }{M}av}.  \label{fseq02}
\end{equation}

Considering $\tilde{r}=\frac{L}{M}v$, follows:

\begin{equation}
I_{T}\left( \tilde{r}\right) =\sqrt{\frac{M}{L}}I_{W}\left( \frac{M}{L}%
\tilde{r}\right) e^{j\frac{2\pi }{L}a\tilde{r}}.  \label{fseq03}
\end{equation}

Since the image is the absolute value of the holographic
representation, it follows:

\begin{equation}
\left| I_{T}\left( \tilde{r}\right) \right| =\sqrt{\frac{M}{L}}\left|
I_{W}\left( \frac{M}{L}\tilde{r}\right) \right| .  \label{fseq04}
\end{equation}

Which proves that, being
$\left|I_{T}\left(\tilde{r}\right)\right|$ a
subsampling of $\sqrt{\frac{M}{L}}\left|I_{W}\left(\frac{M}{L}\tilde{r}%
\right)\right|$, it is a good estimate of $I_{0}$(multiplying $%
\left|I_{T}\left(\tilde{r}\right)\right|$ by $\sqrt{\frac{L}{M}}$), in case
of $I\left(k\right)$ constant.

\section{Continuous Holographic Fourier Transform \label{CHT}}

In this section we extend some developments of section
\ref{sec:DHFT} for continuous signals. Therefore, we start with a
definition for the Continuous Fourier Holographic Transform
(\textbf{CHFT}):

\begin{equation}
H\left( \omega \right) =\int_{-\infty }^{\infty }{f}\left( x\right) \exp
\left( -2\pi j\omega x\right) \exp \left( 2\pi jP\left( x\right) \right) dx,
\label{CHT01}
\end{equation}
where ${P}\left( x\right) $ is a random phase. Following the
development of section \ref{sec:DHFT}, we will also consider a
window in the Fourier domain and check the quality of the recovered
signal. Before this, we shall arrange the terms of the integral in
the following form:

\begin{equation}
H\left( \omega \right) =\int_{-\infty }^{\infty }\exp \left( -2\pi j\omega
x\right) [{f}\left( x\right) \exp \left( 2\pi jP\left( x\right) \right) ]dx.
\label{CHT02}
\end{equation}

We can think about the Fourier transform of the signal $h\left( x\right) ={f}%
\left( x\right) \exp \left( 2\pi jP\left( x\right) \right) ,$ which
is well defined if $f\in L^{2}.$ Now, we use a window function
(low-pass filter) given by expression:

\begin{equation}
W\left( \omega \right) =\left\{
\begin{array}{c}
1,\ \ -k\leq \omega \leq k, \\
0,\quad otherwise.
\end{array}
\right.  \label{CHT03}
\end{equation}

The filtering of signal $h$ by $W$ can be represented, in the
Fourier domain by:

\begin{equation}
G\left( \omega \right) =W\left( \omega \right) H\left( \omega
\right), \label{CHT04}
\end{equation}
where $G\left( \omega \right) $ is the Fourier transform of the
output signal and $H$ is the Fourier transform of $h$, that is, the
CHFT just defined in expression (\ref{CHT01}). The convolution
theorem applied to expression (\ref{CHT04}) renders:

\begin{equation}
g\left( x\right) =\int_{-\infty }^{\infty }h\left( y\right) w\left(
x-y\right) dy,  \label{CHT05}
\end{equation}
which is equivalent to:

\begin{equation}
g\left( x\right) =\int_{-\infty }^{\infty }f\left( y\right) \exp \left( 2\pi
jP\left( y\right) \right) w\left( x-y\right) dy.  \label{CHT06}
\end{equation}

A simple numerical scheme to compute expression (\ref{CHT06}) offers
an interesting application of Lemma 1. In fact, if we approximate
that integral by the Riemann sum:

\begin{equation}
g\left( x\right) \approx \sum_{n=0}^{N}\left[ f\left( y_{n}\right)
w\left( x-y_{n}\right) \Delta y\right] \exp \left( 2\pi jP\left(
y_{n}\right) \right), \label{CHT07}
\end{equation}
we can , from a straightforward application of Lemma 1, estimate the $%
f\left( x\right) $ through:

\begin{equation}
f\left( x\right) \approx \left( \sum_{n=0}^{N}\left[ f\left(
y_{n}\right) w\left( x-y_{n}\right) \Delta y\right] ^{2}\right)
^{1/2}.
\end{equation}

This expression is used in \cite{Gilson-BrunoFig2006} to check the
quality of the recovered signal.

\subsection{CHFT Analysis}

The first point to be demonstrated is the fact that we can recover the input
signal ${f}\left( x\right) $ if we take the inverse Fourier transform of the
CHT given b y:

\begin{equation}
F^{-1}\left( H\left( \omega \right) \right) =\int_{-\infty }^{\infty
}H\left( \omega \right) \exp \left( 2\pi j\omega x\right) d\omega .
\end{equation}

If we substitute the CHT $H\left( \omega \right) $ by expression (\ref{CHT01}%
) we obtain:

\begin{equation}
F^{-1}\left( H\left( \omega \right) \right) =\int_{-\infty }^{\infty }\left[
\int_{-\infty }^{\infty }{f}\left( y\right) \exp \left( -2\pi j\omega
y\right) \exp \left( 2\pi jP\left( y\right) \right) dy\right] \exp \left(
2\pi j\omega x\right) d\omega .
\end{equation}

Using development of \cite{DjairoIMPA} we shall introduce the kernels $%
\exp \left( -\omega ^{2}/n^{2}\right) $ in order to deal with
convergence problems for this integral. Therefore, we get:

\begin{equation}
f_{n}\left( x\right) =\int_{-\infty }^{\infty }\left[ \int_{-\infty
}^{\infty }{f}\left( y\right) \exp \left( -2\pi j\omega y\right) \exp \left(
2\pi jP\left( y\right) \right) dy\right] \exp \left( 2\pi j\omega x\right)
\exp \left( -\frac{\omega ^{2}}{n^{2}}\right) d\omega .
\end{equation}

Now, we can change the order of integration and re-write this
expression as:

\begin{equation}
f_{n}\left( x\right) =\int_{-\infty }^{\infty }{f}\left( y\right) \exp
\left( 2\pi jP\left( y\right) \right) \left[ \int_{-\infty }^{\infty }\exp %
\left[ 2\pi j\omega \left( x-y\right) \right] \exp \left( -\frac{\omega ^{2}%
}{n^{2}}\right) d\omega \right] dy.  \label{fn00}
\end{equation}

The inner integral has analytical solution named by $k_{n}\left(
x-y\right) : $

\begin{equation}
k_{n}\left( x-y\right) =\int_{-\infty }^{\infty }\exp \left[ 2\pi
j\omega \left( x-y\right) \right] \exp \left( -\frac{\omega
^{2}}{n^{2}}\right) d\omega,
\end{equation}
which gives:

\begin{eqnarray}
k_{n}\left( x-y\right) &=&\int_{-\infty }^{\infty }\cos [2\pi j\omega
(x-y)]\exp \left( -\frac{\omega ^{2}}{n^{2}}\right) d\omega + \\
&&j\int_{-\infty }^{\infty }\sin [2\pi j\omega (x-y)]\exp \left( -\frac{%
\omega ^{2}}{n^{2}}\right) d\omega .
\end{eqnarray}

From the fact that $sine$ is an odd function and $exp$ is an even
one, it is easy to shown that the imaginary part will be zero and
the real part of the last expression can be computed as:

\begin{equation}
k_{n}\left( x-y\right) =2\int_{0}^{\infty }\cos [2\pi \omega (x-y)]\exp
\left( -\frac{\omega ^{2}}{n^{2}}\right) d\omega .
\end{equation}

This integral has analytical solution that can be obtained by
expression \cite{Spiegel2004}:

\begin{equation}
\int_{0}^{\infty }\cos bx\exp \left( -ax^{2}\right) dx=\frac{1}{2}\sqrt{%
\frac{\pi }{a}}\exp \left( \frac{-b^{2}}{4a}\right) .
\end{equation}

Henceforth, the application of this expression gives:

\begin{equation}
k_{n}\left( x-y\right) =n\sqrt{\pi }\exp \left[ -\frac{\left[ 2\pi \left(
x-y\right) \right] ^{2}n^{2}}{4}\right] .
\end{equation}

Therefore, returning to equation (\ref{fn00}) we obtain:

\begin{equation}
f_{n}\left( x\right) =\int_{-\infty }^{\infty }{f}\left( y\right) \exp
\left( 2\pi jP\left( y\right) \right) k_{n}\left( x-y\right) dy.
\end{equation}

It can be proved that (see \cite{DjairoIMPA} for details):

\begin{equation}
\lim_{n\rightarrow \infty }f_{n}\left( x\right) ={f}\left( x\right) \exp
\left( 2\pi jP\left( x\right) \right) .  \label{djairo100}
\end{equation}

But, we have also:

\begin{equation}
f_{n}\left( x\right) =\int_{-\infty }^{\infty }H\left( \omega \right) \exp
\left( 2\pi j\omega x\right) \exp \left( -\frac{\omega ^{2}}{n^{2}}\right)
d\omega .
\end{equation}

It is also possible to prove that:

\begin{equation}
\lim_{n\rightarrow \infty }\int_{-\infty }^{\infty }H\left( \omega \right)
\exp \left( 2\pi j\omega x\right) \exp \left( -\frac{\omega ^{2}}{n^{2}}%
\right) d\omega =\int_{-\infty }^{\infty }H\left( \omega \right) \exp \left(
2\pi j\omega x\right) d\omega .  \label{djairo200}
\end{equation}

From (\ref{djairo100}) and (\ref{djairo200}) it follows that:

\begin{equation}
{f}\left( x\right) \exp \left( 2\pi jP\left( x\right) \right) =\int_{-\infty
}^{\infty }H\left( \omega \right) \exp \left( 2\pi j\omega x\right) d\omega ,
\end{equation}
and so, as expected the whole signal is recovered as the amplitude
of the inverse Fourier transform of the CHT.

\section{Progressive Transmission \label{ProgT}}

In this section we propose a progressive transmission approach based
on the DHFT. This is performed by dividing the image (in the
holographic domain) in several equally sized portions. Then, we take
the recovered signal of a portion and incrementally add those ones
corresponding to the other pieces.

Formally, given $W_{1}$ and $W_{2}$ two window functions similar to
expression (\ref{win00}), we have:

\begin{equation}
I_{w_{1}}\left( r\right) =\sum\limits_{u=0}^{M-1}H\left( u\right)
W_{1}\left( u\right) \frac{1}{\sqrt{M}}e^{-j\frac{2\pi }{M}ur},
\label{progress00}
\end{equation}

\begin{equation}
I_{w_{2}}\left( r\right) =\sum\limits_{u=0}^{M-1}H\left( u\right)
W_{2}\left( u\right) \frac{1}{\sqrt{M}}e^{-j\frac{2\pi }{M}ur},
\label{progress01}
\end{equation}
where $H\left( u\right) $ is the DHFT defined in equation
(\ref{holo00}). Thus, the linearity of the Fourier transform allows
to write:

\begin{equation}
I_{w_{1}}\left( r\right) +I_{w_{2}}\left( r\right)
=\sum\limits_{u=0}^{M-1}H\left( u\right) \left( W_{1}\left( u\right)
+W_{2}\left( u\right) \right) \frac{1}{\sqrt{M}}e^{-j\frac{2\pi }{M}ur},
\label{progress02}
\end{equation}
and so:

\begin{equation}
I_{w_{1}}\left( r\right) +I_{w_{2}}\left( r\right)
=I_{_{w_{1}+_{w_{2}}}}\left( r\right) .  \label{win01}
\end{equation}

Observe that if $W_{1}$, $W_{2}$ are disjoint windows, $W_{1}+W_{2}$
is another window function. Consequently, there is not changes in
the spectrum of the signal inside the window.

Thus, following expression (\ref{progress00})-(\ref{progress01}), each recovered portion ($%
I_{w_{1}}\left( r\right) $ and $I_{w_{2}}\left( r\right) $ above) can be
considered as a packet to be transmitted, whose arrival order does not
matter.

\section{Conclusions \label{Concl}}

In this paper we discuss some results concerning to Fourier analysis
and Holographic representation of images.

The approach described in section \ref{ProgT} must be analyzed from
the viewpoint of progressive transmission approaches
\cite{DBLP:journals/tip/KamWG99,DBLP:journals/EG/Chin2004}.
Certainly, the need of double precision representation and the known
instability of Fourier analysis \cite{Chui1992} may be several
limitations for such method, if compared with other ones
\cite{DBLP:journals/EG/Chin2004,DBLP:journals/tip/KamWG99}.
Probably, the main areas of application for the Holographic
transform would be in distributed words and watermarking methods.
Besides a more complete theoretical analysis for CHFT must offered
in order to quantify how good the estimate in expression
(\ref{CHT06}) is for the input signal $f$.

\bibliographystyle{plain}

png
\section{Appendix \label{Appendix}}
In this section we review interesting properties in discrete Fourier
analysis.

\textbf{Property A1:} The circular convolution theorem states that
the DFT of the circular convolution of two sequences is equal to the
product of their DFTs, that is, if:

\begin{equation}
x_{2}\left( n\right) =\sum\limits_{k=0}^{N-1}h\left( n-k\right)
_{c}x_{1}\left( k\right) ,\quad 0\leq n\leq 1,
\label{circul-conv00}
\end{equation}
where $h\left( n-k\right) _{c}$ is the a circular kernel (see
\cite{Jain89}) then:

\begin{equation}
DFT\left[ x_{2}\left( n\right) \right] =DFT\left[ h\left( n\right)
\right] DFT\left[ x_{1}\left( n\right) \right] .  \label{circul00}
\end{equation}

Expression (\ref{circul00}) can be written as:

\begin{equation*}
\sum\limits _{k=0}^{N-1}x_{2}\left(k\right)e^{-j\frac{2\pi}{N}%
vk}=\sum\limits _{k=0}^{N-1}h(k)e^{-j\frac{2\pi}{N}vk}\sum\limits
_{k=0}^{N-1}x_{1}\left(k\right)e^{-j\frac{2\pi}{N}vk}.
\end{equation*}

If we divide this expression by $\sqrt{N}$ we get:

\begin{equation*}
\sum\limits_{k=0}^{N-1}x_{2}\left( k\right) \frac{1}{\sqrt{N}}e^{-j\frac{%
2\pi }{N}vk}=\sum\limits_{k=0}^{N-1}h(k)\frac{1}{\sqrt{N}}e^{-j\frac{2\pi }{N%
}vk}\sum\limits_{k=0}^{N-1}x_{1}\left( k\right) e^{-j\frac{2\pi
}{N}vk}.
\end{equation*}

Therefore, we get the same relation for the Unitary Discrete Fourier
Transform:

\begin{equation}
UDFT\left[ x_{2}\left( n\right) \right] =UDFT\left[ h\left( n\right)
\right] DFT\left[ x_{1}\left( n\right) \right]  \label{circul01}.
\end{equation}

Now, if we define

\begin{equation*}
x_{1}\left( k\right) =\left\{
\begin{array}{l}png
1,\quad k=a, \\
0,\quad k\neq a,
\end{array}
\right.
\end{equation*}
we have that, from expression (\ref{circul-conv00}), $x_{2}\left(
n\right)
=h\left( n-a\right) _{c}$, i.e., $h\left( n\right) $ is a circular shift of $%
x_{2}\left( n\right) $. Hence

\begin{equation*}
UDFT\left[ x_{2}\left( n\right) \right] =UDFT\left[ h\left( n\right)
\right] \sum\limits_{k=0}^{N-1}x_{1}\left( k\right) e^{-j\frac{2\pi
}{N}vk},
\end{equation*}
which implies

\begin{equation*}
UDFT\left[x_{2}\left(n\right)\right]=UDFT\left[h\left(n\right)\right]e^{-j%
\frac{2\pi}{N}va}.png
\end{equation*}

\textbf{Property A2}. Let us consider the summation:

\begin{equation}
\sum\limits_{u=0}^{L-1}e^{-j2\pi xu}=\sum\limits_{u=0}^{L-1}\left(
e^{-j2\pi x}\right) ^{u}.  \label{geom000}
\end{equation}

By remembering the geometric series:
\begin{equation*}
S=\sum\limits_{u=0}^{L-1}a^{u}=1+a+a{{}^{2}}+\cdots +a^{L-1},
\end{equation*}
we know that:
\begin{equation*}
S=\frac{1-a^{L}}{1-a}.
\end{equation*}

If we apply this result in expression (\ref{geom000}) we get:
\begin{equation}
\sum\limits_{u=0}^{L-1}e^{-j2\pi xu}=\frac{1-\left( e^{-j2\pi x}\right) ^{L}%
}{1-e^{-j2\pi x}}=\frac{1-e^{-j2\pi xL}}{1-e^{-j2\pi
x}}=\frac{e^{-j\pi xL}\left( e^{j\pi xL}-e^{-j\pi xL}\right)
}{e^{-j\pi x}\left( e^{j\pi x}-e^{-j\pi x}\right) },  \label{geom01}
\end{equation}

But we know that:
\begin{equation*}
\sin \left( \pi x\right) =\frac{e^{j\pi x}-e^{-j\pi x}}{2j},
\end{equation*}

\begin{equation*}
\sin \left( L\pi x\right) =\frac{e^{j\pi xL}-e^{-j\pi xL}}{2j}.
\end{equation*}
Therefore, expression (\ref{geom01}) becomes:
\begin{equation}
\sum\limits_{u=0}^{L-1}e^{-j2\pi xu}=\frac{e^{-j\pi xL}2j\sin \left(
L\pi x\right) }{e^{-j\pi }2j\sin \left( \pi x\right) }=e^{-j\left(
L-1\right) \pi x}\frac{\sin \left( L\pi x\right) }{\sin \left( \pi
x\right) }. \label{geom02}
\end{equation}

Hence, if $x=0$ we get:

\begin{equation*}
\sum\limits _{u=0}^{L-1}e^{-j2\pi xu}=L.
\end{equation*}

We can also re-write expression (\ref{geom02}) by using the $\mathnormal{sinc%
}$ function, which is defined by: $\mathnormal{sinc}\left(x\right)=\frac{%
\sin\left(x\right)}{x}$.

Thus, that expression becomes:
\begin{equation*}
\sum\limits_{u=0}^{L-1}e^{-j2\pi xu}=e^{-j\left( L-1\right) \pi
x}\frac{L\pi
x\cdot sinc\left( L\pi x\right) }{\pi x\cdot sinc\left( \pi x\right) }%
=Le^{-j\left( L-1\right) \pi x}\frac{sinc\left( L\pi x\right)
}{sinc\left( \pi x\right) }.
\end{equation*}

\end{document}